\pdfoutput=1

\documentclass[11pt]{article}

\usepackage{acl}

\usepackage{times}
\usepackage[scaled=0.86]{helvet}

\usepackage{latexsym}

\usepackage[T1]{fontenc}

\usepackage[utf8]{inputenc}

\usepackage{microtype}

\usepackage{inconsolata}

\usepackage{booktabs}       \usepackage{amsmath,amssymb,amsfonts}
\allowdisplaybreaks

\usepackage{newtxmath}

\usepackage{subcaption}

\usepackage{relsize}
\usepackage{xspace}

\let\tablefont=\scriptsize

\newcommand{\sig}[1]{#1*}

\newcommand\ce{CE\xspace}

\newcommand\ova{MLL\xspace} \newcommand\sr{SR\xspace}
\newcommand\srn{SRN\xspace}
\newcommand\imb{\textup{weighting}}

\newcommand\labels{{\textsf{SUP}}}
\newcommand\labelr{{\textsf{REF}}}
\newcommand\labeln{{\textsf{NEI}}}

\newcommand\numclass{3}

\newcommand\lbl{y}
\newcommand\prob{p}
\newcommand\complbl{\bar{\lbl}}
\newcommand\weight{w}

\newcommand\lblvec{\mathbf{\lbl}}
\newcommand\probvec{\mathbf{\prob}}

\newcommand\lossfunc{L}
\newcommand\regufunc{R}

\title{Rethinking Loss Functions for Fact Verification}

\author{
Yuta Mukobara${}^{\mathsf{a},\dag}$ \quad Yutaro Shigeto${}^{\mathsf{b},\mathsf{c},\ddag}$ \quad Masashi Shimbo${}^{\mathsf{b},\mathsf{c}}$\\
  ${}^{\mathsf{a}}\,$Tokyo Institute of Technology \quad
  ${}^{\mathsf{b}}\,$STAIR Lab, Chiba Institute of Technology \quad
  ${}^{\mathsf{c}}\,$RIKEN AIP\\
   \texttt{mukobara.y.aa@m.titech.ac.jp} \quad \texttt{\{shigeto,shimbo\}@stair.center} \\}

\begin{document}
\maketitle

\begingroup
\def\thefootnote{$\dag$}\footnotetext{Work conducted during an internship at STAIR Lab.}
\def\thefootnote{$\ddag$}\footnotetext{Corresponding author.}
\endgroup

\begin{abstract}
  We explore loss functions for fact verification in the FEVER shared task.
  While the cross-entropy loss is a standard objective for training verdict predictors, it fails to capture the heterogeneity among the FEVER verdict classes.
  In this paper, we develop two task-specific objectives tailored to FEVER. 
  Experimental results confirm that the proposed objective functions outperform the standard cross-entropy.
  Performance is further improved when these objectives are combined with simple class weighting, which effectively overcomes the imbalance in the training data.
  The source code is available.\footnote{\url{https://github.com/yuta-mukobara/RLF-KGAT}}
\end{abstract}

\section{Introduction}
\label{sec:introduction}

The Fact Extraction and VERification (FEVER) shared task \cite{thorne-etal-2018-fever} challenges
systems to verify a given claim by referencing Wikipedia articles. A system for FEVER typically begins by extracting sentences from Wikipedia that potentially support or refute the claim.
Subsequently, the verdict predictor in the system classifies the claim, in conjunction with the retrieved sentences, into one of three verdict classes:
\begin{itemize}
    \item Supported ($\labels$): The retrieved sentences contain evidence supporting the given claim.
    \item Refuted ($\labelr$): The retrieved sentences contain evidence that refutes the claim.
    \item Not Enough Information ($\labeln$): The retrieved sentences do not contain sufficient evidence to support or refute the claim.
\end{itemize}
As this verification step is a multiclass classification task,
verdict predictors are usually trained using the cross-entropy loss function.
However,
cross-entropy treats all misclassification types uniformly, which is problematic
given the heterogeneity among the verdict classes in FEVER;
labels $\labels$ and $\labelr$ both assume evidence is present in the retrieved sentences, whereas a claim is deemed $\labeln$ only when such evidence is missing.
Consequently, it is debatable, for example, whether misclassifying a $\labels$ claim as $\labelr$ or as $\labeln$ should be considered equally severe errors,
especially when the retrieved sentences indeed contain supporting evidence, such as when a verdict predictor is trained with oracle sentences.

In this paper, we explore objective functions designed to capture the heterogeneity among verdict classes.

\paragraph{Notation}

For a $K$-class classification problem,
let $\lblvec = (y_1, \dots , y_K) \in \{0, 1\}^K$ denote a one-hot class representation vector where each index represents a class. Depending on the context, we also use $\lblvec$ to denote the corresponding class itself.
Let $\probvec = (p_1, \dots, p_K) \in [0,1]^K$ denote a predicted class distribution (i.e., $\sum_{i=1}^K p_i = 1$). For FEVER verdict prediction, $K=3$, and let
the indexes $1, 2, 3$ correspond to $\labels, \labelr, \labeln$, respectively.

\section{Proposed Method}
\label{sec:proposed}

\subsection{Cross-entropy Loss Function}
\label{sec:cross-entropy}

We first review the (categorical)  cross-entropy loss, which is a common objective function
for multiclass classification, including FEVER verdict prediction \cite{liu-etal-2020-fine,tymoshenko-moschitti-2021-strong}.

In a $K$-class classification task, the cross-entropy loss for a sample with its one-hot class vector $\lblvec = (y_1 , \dots, y_K)$ is defined as:
\begin{equation}
  \lossfunc_\textup{\ce} (\lblvec, \probvec) = - \sum_{i=1}^{K} \lbl_i \log \prob_i ,
  \label{eq:cross-entropy}
\end{equation}
where $\probvec = (\prob_1, \dots, \prob_K)$  is the class probability distribution derived from the output of a classifier through a softmax function.

\subsection{Loss Functions for Verdict Prediction}
\label{sec:comp-regularizer}

To address the heterogeneity of verdict classes outlined in Section~\ref{sec:introduction},
we implement penalties of varying magnitudes contingent on the type of prediction errors.
To be precise, our objectives impose more severe penalties for incorrectly classifying $\labels$ claims as $\labelr$, or $\labelr$ claims as $\labels$,
considering that classes $\labels$ and $\labelr$ are contradictory when the retrieved sentences contain correct evidence.
Note that this last condition is constantly met during training with oracle sentences in the FEVER dataset.

\subsubsection{Multi-label logistic loss}
\label{sec:one-vs-all}

Before presenting our loss functions for FEVER, we introduce the multi-label logistic (\ova) loss \cite{Baum1988supervised}.
Although this loss is not suited for FEVER verdict prediction,
its inclusion of loss terms for complementary classes helps illustrate our approach.

The \ova loss is defined as the sum of logistic losses (binary cross-entropy) over $K$ components of the predictor's output $\probvec$:
\begin{align}
  \lossfunc_{\textup{\ova}} (\lblvec, \probvec) 
& = - \sum_{i=1}^{K} [ \lbl_i \log \prob_i + \lambda \complbl_i \log (1 - \prob_i) ], \nonumber \\
     & = \lossfunc_{\textup{\ce}}(\lblvec, \probvec) + \lambda \regufunc_{\textup{\ova}}(\lblvec, \probvec) \label{eq:ovr-objective}
\end{align}
where: \begin{equation}
\regufunc_{\textup{\ova}} (\lblvec, \probvec) = - \sum_{i=1}^{K} \complbl_i \log (1 - \prob_i) , \label{eq:ovr-reg-nonexpanded}
\end{equation}
with $\complbl_i = 1 - y_i$.
As Eq.~\eqref{eq:ovr-objective} shows, the \ova loss consists of the primary cross-entropy term and an auxiliary term $\regufunc_{\textup{\ova}}$ for complementary classes.
Also note that, in the original \ova loss, $\lambda=1$, but we treat $\lambda\ge 0$ as a hyperparameter that can also take a different value to control the balance between two terms.

Originally, since the \ova loss was designed for multi-label classification, the $K$ outputs of a predictor are treated as independent variables.
Therefore, each component of the prediction vector $\probvec$ is independently normalized using the sigmoid function.
In contrast, within the scope of this paper, $\probvec$ forms a probability distribution via the softmax function, suitable for a multi-class setting of FEVER.

One interpretation of this loss is that
the predicted class distribution $\probvec = (\prob_1, \dots, \prob_K)$ is viewed not as the outcome of a single $K$-class classification task, but
as the outcomes of $K$ ``one-versus-rest'' binary classification tasks;
in each of these tasks, one of the $K$ classes is treated as the positive class, while the remaining $K-1$ classes are treated collectively as the negative class,
and then individual tasks evaluated by the logistic loss.

\paragraph{Application to verdict prediction}
In Eqs.~\eqref{eq:ovr-objective} and \eqref{eq:ovr-reg-nonexpanded}, $\complbl_i = 1 - \prob_i$ indicates the membership of the $i$th class in the complement of class $\lblvec$, i.e., in the set $Y \setminus \{ \lblvec \}$.
In the context of FEVER, the complement sets for individual verdict classes are
$\overline{\labels} = \{ \labelr, \allowbreak \labeln \}$, $\overline{\labelr} = \{ \labels, \allowbreak \labeln \}$, and $\overline{\labeln} = \{ \labels, \allowbreak \labelr \}$.
Now, setting $K=3$ and recalling that class indexes $1, 2, 3$ represent $\labels, \labelr, \labeln$, respectively, we have:
\begin{align}
  \rlap{$ \displaystyle \regufunc_\textup{\ova} (\lblvec, \probvec)$}
  \quad \nonumber \\
  & = \begin{cases}
        - \log (1 - \prob_2) - \log (1 - \prob_3), &\text{if } \lbl_1 = 1, \\
        - \log (1 - \prob_3) - \log (1 - \prob_1), &\text{if } \lbl_2 = 1, \\
        - \log (1 - \prob_1) - \log (1 - \prob_2), &\text{if } \lbl_3 = 1.
      \end{cases} 
    \label{eq:ovr-reg}
\end{align}

Eq.~\eqref{eq:ovr-reg} is symmetric over classes,
which shows that the \ova loss does not account for the heterogeneity among verdict classes, much like the cross-entropy loss.
Later experiments in Section~\ref{sec:experiment} indeed demonstrate that the \ova loss does not improve over the standard cross-entropy in terms of prediction accuracy.

\subsubsection{Reducing penalties for false \labeln}
\label{sec:srn}

We address the issue of heterogeneous verdict classes by modifying the composition of complement sets in the \ova loss. 

Specifically, in our first FEVER-specific loss function,
we treat classes $\labels$ and $\labelr$ as their sole complementary class, excluding $\labeln$. To be precise, we let $\overline{\labels} = \{ \labelr \}$,  $\overline{\labelr} = \{ \labels \}$, whereas $\overline{\labeln} = \{\labels, \labelr \}$ is unchanged.
Accordingly, the membership indicator $\complbl_i$ is changed to:
\begin{align}
  \complbl^{\,\textup{\smaller\srn}}_i =
  \begin{cases}
    1 - \lbl_i, & \text{if } i = 1, 2, \\
    0, & \text{if } i = 3 ,
  \end{cases}
  \label{eq:srn-y-comp}
\end{align}
which results in: \begin{align}
  \rlap{$ \displaystyle \regufunc_{\textup{\srn}} (\lblvec, \probvec) = - \sum_{i=1}^3 \complbl^{\,\textup{\smaller\srn}} _i \log (1 - \prob_i) $} \quad \nonumber \\[-1ex]
  &= - \sum_{i=1}^2 (1 - \lbl_i) \log (1 - \prob_i) \nonumber \\
&=
  \begin{cases}
  - \log (1 - \prob_2), &\text{if }  \lbl_1 = 1, \\
  - \log (1 - \prob_1), &\text{if } \lbl_2 = 1, \\
  - \log (1 - \prob_1)  - \log (1 - \prob_2), &\text{if } \lbl_3 = 1.
  \end{cases} 
  \label{eq:srn-regularizer}
\end{align}
Comparing the last formula with Eq.~\eqref{eq:ovr-reg}, we see that
$\regufunc_{\textup{\srn}}$ effectively reduces penalties for misclassifying $\labels$ or $\labelr$ claims (i.e., $\lbl_1 = 1$ or $\lbl_2 = 1$) as $\labeln$.
Combining the auxiliary loss with the cross entropy loss, we obtain the overall objective:
\begin{align}
  \rlap{$\displaystyle \lossfunc_\textup{\srn} (\lblvec, \probvec) = \lossfunc_\textup{\ce} (\lblvec, \probvec) + \lambda \regufunc_\textup{\srn} (\lblvec, \probvec) $} \quad \nonumber \\
& = - \sum_{i=1}^{\numclass} \lbl_i \log \prob_i - \lambda \sum_{i=1}^2 (1 - \lbl_i) \log (1 - \prob_i) .
    \label{eq:srn-objective}
\end{align}

\subsubsection{Exclusive penalties for $\labels$/$\labelr$ confusion}
\label{sec:sr}

Alternatively,
we can define an auxiliary loss focusing only on the contradictory nature of $\labels$ and $\labelr$ and disregarding $\labeln$ entirely.
To this end, we define $\overline{\labeln} = \emptyset$.
For $\labels$ and $\labelr$, their complementary sets are defined in the same way as the \srn loss term,
namely,
$\overline{\labels} = \{ \labelr \}$ and $\overline{\labelr} = \{ \labels \}$.
The corresponding membership indicator is given by:
\begin{align*}
  \complbl^{\,\textup{\smaller\sr}}_i =
  \begin{cases}
    (1 - \lbl_i) (1 - \lbl_3), & \text{if } i = 1, 2, \\
    0, & \text{if } i = 3.
  \end{cases}
\end{align*}
The newly introduced factor $(1 - \lbl_3)$ ensures $\complbl^{\,\textup{\smaller\sr}}_i$ remains $0$ when the gold label is $\labeln$ (and thus $y_3 = 1$).
This produces our second auxiliary loss function for FEVER:
\begin{align}
  \regufunc_\textup{\sr} (\lblvec, \probvec)
                        & = - \sum_{i=1}^3 \complbl^{\,\textup{\smaller\sr}} _i \log (1 - \prob_i) \nonumber \\[-1.2ex]
                        & =  - (1-\lbl_3) \sum_{i=1}^2 (1-\lbl_i) \log (1 - \prob_i) \nonumber               \\
                        & = \begin{cases}
                              - \log (1 - \prob_2), & \text{if }  \lbl_1 = 1,                                \\
                              - \log (1 - \prob_1), & \text{if } \lbl_2 = 1,                                 \\
                              0,                    & \text{if } \lbl_3 = 1.
                            \end{cases}
                          \label{eq:sr-regularizer}
\end{align}
In this loss term, any misclassification involving label $\labeln$ is disregarded;
$\regufunc_{\textup{\sr}}$ imposes no penalty for prediction errors on $\labeln$ claims, nor for misclassifying $\labels$ and $\labelr$ claims as $\labeln$.

The overall objective function, combining $\regufunc_{\textup{\sr}}$ with $\lossfunc_{\textup{\ce}}$, is given as follows:
\begin{align}
  \rlap{$\displaystyle \lossfunc_\textup{\sr} (\lblvec, \probvec) 
  = 
  \lossfunc_{\textup{\ce}} (\lblvec, \probvec) +  \lambda \regufunc_{\textup{\sr}} (\lblvec, \probvec) $} \nonumber \\
  & =
    - \sum_{i=1}^{\numclass} \lbl_i \log \prob_i \nonumber \\[-1.4ex]
  & \qquad
    - \lambda (1-\lbl_3) \sum_{i=1}^2 (1-\lbl_i) \log (1 - \prob_i).
  \label{eq:sr-objective}
\end{align}

\subsection{Class Imbalanced Learning}
\label{sec:imbalanced-learning}

Another non-negligible issue in verdict prediction is the imbalanced training data in the FEVER dataset, 
whose class frequency is shown in Table~\ref{tab:dataset}. 

A popular approach to class imbalance problems \cite{10105457,chawla2002smote}
is class weighting \cite{ren2018learning,cui2019class}, where each term in the objective function is assigned a different weight depending on the class it is associated with.

For example, after weighting applied, the \srn objective in Eq.~\eqref{eq:srn-objective} becomes:
\begin{align}
  \rlap{$\displaystyle \lossfunc_{\textup{\srn}+\imb} (\lblvec, \probvec) $}\;\; \nonumber \\
  &=
  - \sum_{i=1}^{\numclass} \weight_i \left[ \lbl_i \log \prob_i 
  + \complbl_i^{\,\textup{\smaller\srn}} \log (1 - \prob_i) \right],
  \label{eq:srn-imb-objective}
\end{align}
where $w_1$, $w_2$, and $w_3$ are the fixed class weights. The same weighting scheme can be applied to \sr and \ova objective functions; see
Appendix~\ref{sec:summary-objective}.

In our experiments in Section~\ref{sec:experiment}, we use the class-balanced weights of \citet{cui2019class}.
They define
the weight for the $i$th class as:
\begin{align}
  \weight_i = \frac{1- \beta}{1 - \beta^{n_i}}, 
  \label{eq:balanced-weight}
\end{align}
where $n_i$ is the number of training samples in the $i$th class and $\beta$ is a hyperparameter. 
Setting $\beta = 0$ results in uniform weights $\weight_1 = \weight_2 = \weight_3 = 1$, which reduces Eq.~\eqref{eq:srn-imb-objective} to the unweighted one in Eq.~\eqref{eq:srn-objective}.
As $\beta\to 1$, the weights approach the inverse class frequency $1/n_i$.

\section{Experiments}
\label{sec:experiment}

Due to limited space, only the main experimental results are presented below. 
Additional results and analysis can be found in Appendix~\ref{sec:additional-results}.

\subsection{Setups}
\label{sec:experiment-setup}

\begin{table}[t]
\centering
\tablefont
\begin{tabular}{l rrr}
\toprule 
Split & \#$\labels$ & \#$\labelr$ & \#$\labeln$ \\
\midrule 
Train & 80,035      & 29,775      & 35,639      \\
Dev   & 6,666       & 6,666       & 6,666       \\
Test  & 6,666       & 6,666       & 6,666       \\
\bottomrule
\end{tabular}
\caption{
Number of samples (claim-evidence pairs) in the FEVER 2018 dataset.
}
\label{tab:dataset}
\end{table}

\paragraph{Dataset and evaluation criteria}
The FEVER 2018 dataset \cite{thorne-etal-2018-fever} consists of 185,445 claims (Table~\ref{tab:dataset}).
Each claim is assigned a gold class labels, $\labels$, $\labelr$, or $\labeln$.
The gold labels for the test set are not disclosed.

Models are evaluated by prediction label accuracy (LA) and FEVER score (FS).
LA is a standard evaluation criterion for multiclass classification where classification accuracy is computed without considering the correctness of the retrieved evidence.
In FS, a prediction is deemed correct only if the predicted label is correct and the correct evidence is retrieved (in the case of $\labels$ and $\labelr$ claims).
The scores for the test set, for which the gold labels are not disclosed, are computed on the official FEVER scoring site.

\paragraph{Compared models and hyperparameters}

We use KGAT\footnote{\url{https://github.com/thunlp/KernelGAT}} \cite{liu-etal-2020-fine} for both evidence retrieval and verdict prediction.
Multiple prediction models are trained, each with a different objective function.
The objectives employed are:
\begin{itemize}
\item \ce: The cross-entropy loss of Eq.~\eqref{eq:cross-entropy}. 
  This is the standard objective function for FEVER\textsf{}.
  It is used by the original KGAT, and is the baseline in our experiments. 
\item
  \ova:
The multi-label logistic loss of Eq.~\eqref{eq:ovr-objective}.
  As our proposed objectives can be considered its modifications, it is included as another baseline in this comparative study.

\item
  \srn:
  Our first proposed objective (Eq.~\eqref{eq:srn-objective}), which combines the cross-entropy loss with the $\regufunc_{\textup{\srn}}$ auxiliary loss.
\item
  \sr:
  Our second proposed objective (Eq.~\eqref{eq:sr-objective}), which augments the cross-entropy loss with the $\regufunc_{\textup{\sr}}$ auxiliary loss.

\end{itemize} 
Each objective is assessed with and without the class weighting scheme of Eq.~\eqref{eq:balanced-weight}.
A summary of all objective functions evaluated can be found in Appendix~\ref{sec:summary-objective}.
Additionally, all objectives are evaluated with three different backbone networks: BERT Base, BERT Large \cite{devlin-etal-2019-bert}, and RoBERTa Large \cite{liu2019roberta}.

Hyperparameters $\lambda$ in Eqs.~\eqref{eq:ovr-objective},~\eqref{eq:srn-objective},~and \eqref{eq:sr-objective}, and $\beta$ in Eq.~\eqref{eq:balanced-weight} are tuned on the development set.
For other hyperparameters (e.g., learning rate and batch size), the default values set in the KGAT implementation are used. 
Each model is trained three times and the one achieving the highest LA on the development set is selected for evaluation.

\subsection{Results}
\label{sec:experiment-result}

\begin{table}[t]
  \tablefont
  \centering
  \begin{tabular}{lc ll}
    \toprule 
Objective function         & Weighting   & \multicolumn1c{LA}     & \multicolumn1c{FS}            \\
    \midrule                                                                                          
\multicolumn4l{Backbone: BERT Base}                                                               \\
    \cmidrule(lr){1-2}                                                                                 
    {\ce} (baseline)           & --          & 77.81                        & 75.75                   \\
    {\ce}                      & yes         & 78.08 (+0.27)                & 76.02 (+0.27)          \\[0.5ex]
{\ova} ($\lambda$=0.0625)  & --          & 77.84 (+0.03)                & 75.65 (-0.10)           \\
    {\ova} ($\lambda$=0.125)   & yes         & 78.13 (+0.32)                & \textbf{76.06 (+0.31)}  \\[0.5ex]
{\srn} ($\lambda$=0.0625)  & --          & 77.84 (+0.03)                & 75.70 (-0.05)           \\
    {\srn} ($\lambda$=0.0625)  & yes         & 77.83 (+0.02)                & 75.79 (+0.04)           \\[0.5ex]
{\sr} ($\lambda$=0.0625)   & --          & 78.16 (+0.35)                & 75.87  (+0.12)          \\
    {\sr} ($\lambda$=0.25)     & yes         & \sig{\textbf{78.29 (+0.48)}} & \textbf{76.06 (+0.31)}  \\
    \midrule                                                                                          
\multicolumn4l{Backbone: BERT Large}                                                              \\
    \cmidrule(lr){1-2}                                                                                 
{\ce} (baseline)           & --          & 78.20                        & 75.98                   \\
    {\ce}                      & yes         & \sig{78.85 (+0.65)}          & 76.74 (+0.76)           \\[0.5ex]
{\ova} ($\lambda$=0.25)    & --          & \sig{78.94 (+0.74)}          & 76.78 (+0.80)           \\
    {\ova} ($\lambda$=0.03125) & yes         & \sig{78.85 (+0.65)}          & 76.74 (+0.76)           \\[0.5ex]
{\srn} ($\lambda$=0.125)   & --          & \sig{78.68 (+0.48)}          & 76.57 (+0.59)           \\
    {\srn} ($\lambda$=0.25)    & yes         & \sig{78.83 (+0.63)}          & 76.71 (+0.73)           \\[0.5ex]
{\sr} ($\lambda$=0.25)     & --          & \sig{79.02 (+0.82)}          & 76.86 (+0.88)           \\
    {\sr} ($\lambda$=0.125)    & yes         & \sig{\textbf{79.19 (+0.99)}} & \textbf{77.01 (+1.03)}  \\
    \midrule                                                                                          
\multicolumn4l{Backbone: RoBERTa Large}                                                           \\
    \cmidrule(lr){1-2}
{\ce} (baseline)            & --          & 80.19                        & 78.03                  \\
    {\ce}                       & yes         & 80.55 (+0.36)                & 78.54 (+0.51)          \\[0.5ex]
{\ova} ($\lambda$=0.0625)   & --          & 80.00 (-0.19)                & 77.88 (-0.15)          \\
    {\ova} ($\lambda$=0.0625)   & yes         & \sig{80.62 (+0.43)}          & 78.55 (+0.52)          \\[0.5ex]
{\srn} ($\lambda$=0.03125)  & --          & 80.24 (+0.05)                & 78.18 (+0.15)          \\
    {\srn} ($\lambda$=0.03125)  & yes         & \sig{\textbf{80.73 (+0.54)}} & 78.56 (+0.53)          \\[0.5ex]
{\sr} ($\lambda$=0.0625)    & --          & 80.41 (+0.22)                & 78.19 (+0.16)          \\
    {\sr} ($\lambda$=0.03125)   & yes         & \sig{80.70 (+0.51)}          & \textbf{78.63 (+0.60)} \\
    \bottomrule
  \end{tabular}
  \caption{
    Label accuracy (LA) and FEVER score (FS) of KGAT models on the development set, using different loss functions and backbones.
For class-balanced weighting, $\beta$ is set to 0.999999 in all cases.
    The parenthesized figures after LA indicate differences from the baseline cross-entropy loss (\ce) without class-balanced weighting. Asterisks (*) denote the change in prediction from \ce (baseline) is statistically significant ($p<0.05$), as determined by the McNemar test \cite{Mcnemar1947NoteOT}.
  }
  \label{tab:res-ablation}
\end{table}

 \paragraph{Effectiveness of the proposed objective functions}
Table~\ref{tab:res-ablation} shows the results. Trends observed are:
(i) The imbalance weighting consistently improves both LA and FS. 
(ii) The proposed \srn and \sr losses enhance LA in all cases and FS in most cases.
(iii) The simultaneous use of the class-balance weighting and the proposed losses further improves the performance.

Of the two proposed loss types, {\sr} achieves higher scores across all backbone architectures, with the exception of the LA score with RoBERTa Large. Even in the latter case, the difference is marginal (0.03).
For \sr with weighting, the change in predictions from \ce (baseline) is statistically significant irrespective of the backbones. The same is true for \srn with weighting, except when it is used with BERT Base.

Although the \ova loss explicitly has the additional penalty term for the complement sets, it does not account for the label heterogeneity as in the cross-entropy loss (see Section~\ref{sec:one-vs-all}). 
Indeed, there is little difference in the results between \ce and \ova, excluding the BERT Large backbone without weighting.

\begin{table}[t]
\tablefont
\centering
\begin{tabular}{l @{\hspace{0.5em}} rr rr}
\toprule 
                                                & \multicolumn{2}{c}{Dev} & \multicolumn{2}{c}{Test}                                     \\
\cmidrule(lr){2-3} \cmidrule(lr){4-5}
Method                                          & \multicolumn1c{LA}      & \multicolumn1c{FS} & \multicolumn1c{LA} & \multicolumn1c{FS} \\
\midrule 
Backbone: BERT Base                                                                                                                      \\
\cmidrule(lr){1-1}
KGAT~\cite{liu-etal-2020-fine}                  & 78.02                   & 75.88              & 72.81              & 69.40              \\
KGAT (reproduced)                               & 77.81                   & 75.75              & 73.01              & 69.29              \\
KGAT + \sr{} + weighting                        & \textbf{78.29}          & \textbf{76.06}     & \textbf{73.44}     & \textbf{69.88}     \\
\midrule
Backbone: BERT Large                                                                                                                     \\
\cmidrule(lr){1-1}
KGAT~\cite{liu-etal-2020-fine}                  & 77.91                   & 75.86              & 73.61              & 70.24              \\
KGAT (reproduced)                               & 78.20                   & 75.98              & 73.66              & 70.06              \\
KGAT + \sr{} + weighting                        & \textbf{79.19}          & \textbf{77.01}     & \textbf{73.97}     & \textbf{70.71}     \\
\midrule
Backbone: RoBERTa Large                                                                                                                  \\
\cmidrule(lr){1-1}
KGAT~\cite{liu-etal-2020-fine}                  & 78.29                   & 76.11              & 74.07              & 70.38              \\
KGAT (reproduced)                               & 80.19                   & 78.03              & 75.40              & 72.04              \\
KGAT + \sr{} + weighting                        & \textbf{80.70}          & \textbf{78.63}     & \textbf{75.72}     & \textbf{72.53}     \\
\midrule
Non-KGAT SOTA Methods                                                                                                                          \\
\cmidrule(lr){1-1}
Stammbach~\cite{stammbach-2021-evidence}        & --                      & --                 & 79.20              & 76.80              \\
LisT5~\cite{jiang-etal-2021-exploring-listwise} & 81.26                   & 77.75              & 79.35              & 75.87              \\
ProoFVer~\cite{krishna-etal-2022-proofver}      & 80.74                   & 79.07              & 79.47              & 76.82              \\
BEVERS~\cite{dehaven-scott-2023-bevers}         & --                      & --                 & \textbf{80.24}     & \textbf{77.70}     \\
\bottomrule
\end{tabular}
\caption{
Label accuracy (LA) and FEVER score (FS) on the development (Dev) and test sets.
The bold values indicate the best performer in the group.
}
\label{tab:res-best}
\end{table}

 \paragraph{Comparison with SOTA models}
As KGAT with the proposed \sr objective and class-balanced weighting showed consistent performance on the development set, we submit its predictions on the test set to the FEVER scoring site.
Table~\ref{tab:res-best} presents the results, along with those of the original KGAT and state-of-the-art (SOTA) FEVER models.
The proposed methods (KGAT + {\sr} + weighting) consistently outperform the original KGAT (using the standard \ce{} loss) on the test set as well, regardless of the backbone architecture.
These results suggest that the cross-entropy objective is not necessarily optimal for the FEVER task, and our approach offers a means of improvement. 

The scores of KGAT models, including our proposed approach, are lower than those of the SOTA models \cite{stammbach-2021-evidence,jiang-etal-2021-exploring-listwise,krishna-etal-2022-proofver,dehaven-scott-2023-bevers}. 
However, it should be noted that these models owe their better performance in part to the improved retrievers and backbones they use.
Indeed, \citet[Table~12]{dehaven-scott-2023-bevers} report an LA of 76.60 and an FS of 73.21 on the test set, when their BEVERS model is used in combination with the KGAT retriever and the RoBERTa Large backbone.
These figures represent a notable regression from those presented in Table~\ref{tab:res-best},
consequently reducing the advantage over our model (with a test LA of 75.72, and a test FS of 72.53) to less than a 1-point.

\section{Related Work}
\label{sec:related}

The FEVER shared tasks~\cite{thorne-etal-2018-fever,thorne-etal-2019-fever2,aly-etal-2021-fact,aly2021feverous} have been the subject of extensive research.
Most proposed approaches utilize Transformer-based models to embed claims and evidence \cite{tymoshenko-moschitti-2021-strong,jiang-etal-2021-exploring-listwise,stammbach-2021-evidence,dehaven-scott-2023-bevers},
whereas some researchers \cite{zhou-etal-2019-gear,liu-etal-2020-fine} use graph-based methods to aggregate information from multiple pieces of evidence.
None of these studies focus on the objective function to optimize,
and most employ the standard cross-entropy objective.

Recently, \citet{dehaven-scott-2023-bevers} have used class weighting to mitigate class imbalance in the FEVER dataset,
although the detailed weighting scheme is not reported.

In machine learning,
\citet{zhang2004statistical} analyzes various loss functions used for multiclass classification,
including a general form of one-versus-rest (or one-versus-all) loss functions, which also have terms accounting for the complement set of the ground-truth class.
\citet{ishida2017complementary} study complementary-label learning scenarios \cite{ishida2017complementary,yu2018learning,pmlr-v97-ishida19a} extending \citeauthor{zhang2004statistical}'s losses.

\section{Conclusion}
\label{sec:conlclusion}

We introduced loss functions that take into account the heterogeneity of verdict classes in the FEVER task.
In empirical evaluation, they consistently outperformed the standard cross-entropy loss.

In future work, we will evaluate the proposed loss functions in other fact verification tasks. 
We also plan to apply them to SOTA models for FEVER.
As these models use the cross-entropy loss, our auxiliary loss terms are readily applicable.

\section*{Limitations}

Our empirical evaluation was conducted in limited situations.
\begin{itemize}
\item Task (dataset):
  Although our approach proved effective in the FEVER task and dataset \cite{thorne-etal-2018-fever},
  whether it works equally well in other similar tasks and datasets remains unverified.

\item Verdict predictor:
  The effectiveness of our approach was demonstrated only in combination with KGAT~\cite{liu-etal-2020-fine}, a popular prediction model frequently used for benchmarking FEVER methods.
  Being model-agnostic, our loss functions need to be evaluated in combination with more recent models that optimize the cross-entropy loss.
\end{itemize}

\section*{Acknowledgments}

We are grateful to anonymous reviewers for their constructive comments.
This work is partially supported by JSPS Kakenhi Grant 21K17811 to YS.

\appendix

\section{Summary of Objective Functions}
\label{sec:summary-objective}

In the following, we list the formulas for the objective functions used in our experiments.

\paragraph{Cross-entropy objective}
The cross-entropy objective presented in Eq.~\eqref{eq:cross-entropy} is repeated here for convenience.
\begin{equation*}
  \lossfunc_\textup{\ce} (\lblvec, \probvec) = - \sum_{i=1}^{\numclass} \lbl_i \log \prob_i .
\end{equation*}
Its class-weighted version is:
\begin{equation*}
  \lossfunc_{\textup{\ce}+\imb} (\lblvec, \probvec) = - \sum_{i=1}^{\numclass} \weight_i \lbl_i \log \prob_i .
\end{equation*}

\paragraph{{\ova} objective}

The \ova objective of Eq.~\eqref{eq:ovr-objective} is:
\begin{align*}
  \rlap{$\displaystyle \lossfunc_\textup{\ova} (\lblvec, \probvec) 
  = \lossfunc_\textup{\ce} (\lblvec, \probvec) +  \lambda \regufunc_\textup{\ova} (\lblvec, \probvec) $}\quad \nonumber \\
& =  
    - \sum_{i=1}^{\numclass} \left[ \lbl_i \log \prob_i
    + \lambda (1 - \lbl_i) \log (1 - \prob_i) \right],
\end{align*}
and its weighted version is:
\begin{align*}
  \rlap{$\displaystyle \lossfunc_{\textup{\ova}+\imb} (\lblvec, \probvec) 
$}\quad \nonumber \\
& =  
    - \sum_{i=1}^{\numclass} w_i \left[ \lbl_i \log \prob_i
    + \lambda (1 - \lbl_i) \log (1 - \prob_i) \right].
\end{align*}

\paragraph{{\srn} objective}

The \srn objective $\lossfunc_{\textup{\srn}}$, originally presented in Eq.~\eqref{eq:srn-objective}, is restated below, accompanied by its instantiation for individual gold classes: \begin{align*}
  \rlap{$ \displaystyle \lossfunc_\textup{\srn} (\lblvec, \probvec)
  = 
  \lossfunc_\textup{\ce} (\lblvec, \probvec) +  \lambda \regufunc_\textup{\srn} (\lblvec, \probvec) $} \quad \nonumber \\
& =  
    - \sum_{i=1}^{\numclass} \lbl_i \log \prob_i
    - \lambda \sum_{i=1}^2 (1 - \lbl_i) \log (1 - \prob_i)  \nonumber \\
  & =  
    \begin{cases}
      - \log \prob_1 - \log (1 - \prob_2)                  , & \text{if } \lbl_1 = 1, \\
      - \log \prob_2 - \log (1 - \prob_1)                  , & \text{if } \lbl_2 = 1, \\
      - \log \prob_3
      \begin{aligned}[t]
        &  - \log (1 - \prob_1) \\[0.3ex]
        & \qquad - \log (1 - \prob_2) ,   
      \end{aligned} 
      & \raisebox{-3.1ex}{$\displaystyle \text{if } \lbl_3 = 1 .$}
    \end{cases} 
\end{align*}
With class weighting, the objective becomes Eq.~\eqref{eq:srn-imb-objective}, as shown in Section~\ref{sec:comp-regularizer}.
The corresponding expressions for individual gold classes are as follows:
\begin{align*}
  \rlap{$ \displaystyle \lossfunc_{\textup{\srn}+\imb} (\lblvec, \probvec) $} \quad \nonumber \\
&=  
  \begin{cases}
    -\weight_1 \left[ \log \prob_1 + \log (1 - \prob_2)                      \right], & \text{if } \lbl_1 = 1, \\
    -\weight_2 \left[ \log \prob_2 + \log (1 - \prob_1)                      \right], & \text{if } \lbl_2 = 1, \\
    -\weight_3 
    \begin{aligned}[t]
      & \left[ \log \prob_3 + \log (1 - \prob_1) \right. \\[0.3ex]
      & \qquad\quad \left. + \log (1 - \prob_2) \right],   
    \end{aligned}
                                                                                      & \raisebox{-3.1ex}{$\displaystyle \text{if } \lbl_3 = 1 .$}
  \end{cases} 
\end{align*}

\paragraph{{\sr} objective}

The objective $\lossfunc_\textup{\sr}$ is shown below:
\begin{align*}
  \rlap{$\displaystyle \lossfunc_\textup{\sr} (\lblvec, \probvec) 
  = 
  \lossfunc_{\textup{\ce}} (\lblvec, \probvec) +  \lambda \regufunc_{\textup{\sr}} (\lblvec, \probvec) $} \nonumber \\
  & =
    - \sum_{i=1}^{\numclass} \lbl_i \log \prob_i \nonumber \\[-1.1ex]
  & \qquad
    - \lambda (1-\lbl_3) \sum_{i=1}^2 (1-\lbl_i) \log (1 - \prob_i) \nonumber \\
  &=  
  \begin{cases}
  - \log \prob_1 - \log (1 - \prob_2), & \text{if } \lbl_1 = 1, \\
  - \log \prob_2 - \log (1 - \prob_1), & \text{if } \lbl_2 = 1, \\
  - \log \prob_3,                      & \text{if } \lbl_3 = 1.
  \end{cases}
\end{align*}
And the weighted version is:
\begin{align*}
\rlap{$ \displaystyle \lossfunc_{\textup{\sr}+\imb} (\lblvec, \probvec) $} \quad \nonumber \\
  &=  
  \begin{cases}
  -\weight_1 \left[ \log \prob_1 + \log (1 - \prob_2) \right], &\text{if } \lbl_1 = 1, \\
  -\weight_2 \left[ \log \prob_2 + \log (1 - \prob_1) \right], &\text{if } \lbl_2 = 1, \\
  -\weight_3        \log \prob_3, &\text{if } \lbl_3 = 1.
  \end{cases} 
  \label{eq:app:sr-imb-objective}
\end{align*}

\section{Additional Experimental Results}
\label{sec:additional-results}

\subsection{Confusion Matrices}
\label{sec:confusion-matrix-appendix}

\begin{table*}[t]
  \centering
  \tablefont
  \begin{subcaptionblock}[T]{0.48\linewidth}
    \centering
    \begin{tabular}{ll rrr}
      \toprule
                 &           & \multicolumn3c{Prediction}        \\
      \cmidrule{3-5}
                 &           & $\labels$ & $\labelr$ & $\labeln$ \\
      \midrule 
                 & $\labels$ & 5976      & 222       & 468       \\
      Gold       & $\labelr$ & 470       & 5153      & 1043      \\
                 & $\labeln$ & 1051      & 1184      & 4431      \\
      \midrule
           Total &           & 7497      & 6559      & 5942      \\
      \bottomrule
    \end{tabular}
    \caption{Loss = {\ce}, Weighting = no (FS=75.75, LA=77.81)}
  \end{subcaptionblock}
\begin{subcaptionblock}[T]{0.48\linewidth}
    \centering
    \begin{tabular}{ll rrr}
      \toprule 
                 &           & \multicolumn3c{Prediction}        \\
      \cmidrule{3-5}
                 &           & $\labels$ & $\labelr$ & $\labeln$ \\
      \midrule 
                 & $\labels$ & 5862      & 214       & 590       \\
      Gold       & $\labelr$ & 427       & 4906      & 1333      \\
                 & $\labeln$ & 922       & 897       & 4847      \\
      \midrule
           Total &           & 7211      & 6017      & 6770      \\
      \bottomrule
    \end{tabular}
    \caption{Loss = {\ce}, Weighting = yes (FS=76.02, LA=78.08)}
  \end{subcaptionblock}
                                                                 \\[5ex]
  
  \begin{subcaptionblock}[T]{0.48\linewidth}
    \centering
    \begin{tabular}{ll rrr}
      \toprule 
                 &           & \multicolumn3c{Prediction}        \\
      \cmidrule{3-5}
                 &           & $\labels$ & $\labelr$ & $\labeln$ \\
      \midrule 
                 & $\labels$ & 5976      & 201       & 489       \\
      Gold       & $\labelr$ & 510       & 4981      & 1175      \\
                 & $\labeln$ & 1066      & 991       & 4609      \\
      \midrule
           Total &           & 7552      & 6173      & 6273      \\
      \bottomrule
    \end{tabular}
    \caption{Loss = {\ova}, Weighting = no (FS=75.65, LA=77.84)}
  \end{subcaptionblock}
\begin{subcaptionblock}[T]{0.48\linewidth}
    \centering
    \begin{tabular}{ll rrr}
      \toprule 
                 &           & \multicolumn3c{Prediction}        \\
      \cmidrule{3-5}
                 &           & $\labels$ & $\labelr$ & $\labeln$ \\
      \midrule 
                 & $\labels$ & 5785      & 303       & 578       \\
      Gold       & $\labelr$ & 372       & 5098      & 1196      \\
                 & $\labeln$ & 845       & 1079      & 4742      \\
      \midrule
           Total &           & 7002      & 6480      & 6516      \\
      \bottomrule
    \end{tabular}
    \caption{Loss = {\ova}, Weighting = yes (FS=76.06, LA=78.13)}
  \end{subcaptionblock}
                                                                 \\[5ex]
  
  \begin{subcaptionblock}[T]{0.48\linewidth}
    \centering
    \begin{tabular}{ll rrr}
      \toprule 
                 &           & \multicolumn3c{Prediction}        \\
      \cmidrule{3-5}
                 &           & $\labels$ & $\labelr$ & $\labeln$ \\
      \midrule 
                 & $\labels$ & 5919      & 196       & 551       \\
      Gold       & $\labelr$ & 455       & 4876      & 1335      \\
                 & $\labeln$ & 1001      & 894       & 4771      \\
      \midrule
           Total &           & 7375      & 5966      & 6657      \\
      \bottomrule
    \end{tabular}
    \caption{Loss = {\srn}, Weighting = no (FS=75.70, LA=77.84)}
  \end{subcaptionblock}
\begin{subcaptionblock}[T]{0.48\linewidth}
    \centering
    \begin{tabular}{ll rrr}
      \toprule 
                 &           & \multicolumn3c{Prediction}        \\
      \cmidrule{3-5}
                 &           & $\labels$ & $\labelr$ & $\labeln$ \\
      \midrule 
                 & $\labels$ & 5766      & 239       & 661       \\
      Gold       & $\labelr$ & 444       & 4958      & 1264      \\
                 & $\labeln$ & 864       & 962       & 4840      \\
      \midrule
           Total &           & 7074      & 6159      & 6765      \\
      \bottomrule
    \end{tabular}
    \caption{Loss = {\srn}, Weighting = yes (FS=75.79, LA=77.83)}
  \end{subcaptionblock}
                                                                 \\[5ex]
  
  \begin{subcaptionblock}[T]{0.48\linewidth}
    \centering
    \begin{tabular}{ll rrr}
      \toprule 
                 &           & \multicolumn3c{Prediction}        \\
      \cmidrule{3-5}
                 &           & $\labels$ & $\labelr$ & $\labeln$ \\
      \midrule 
                 & $\labels$ & 5948      & 221       & 497       \\
      Gold       & $\labelr$ & 461       & 4969      & 1236      \\
                 & $\labeln$ & 1014      & 939       & 4713      \\
      \midrule
           Total &           & 7423      & 6129      & 6446      \\
      \bottomrule
    \end{tabular}
    \caption{Loss = {\sr}, Weighting = no (FS=75.87, LA=78.16)}
  \end{subcaptionblock}
\begin{subcaptionblock}[T]{0.48\linewidth}
    \centering
    \begin{tabular}{ll rrr}
      \toprule 
                 &           & \multicolumn3c{Prediction}        \\
      \cmidrule{3-5}
                 &           & $\labels$ & $\labelr$ & $\labeln$ \\
      \midrule 
                 & $\labels$ & 5979      & 228       & 459       \\
      Gold       & $\labelr$ & 457       & 5031      & 1178      \\
                 & $\labeln$ & 1080      & 939       & 4647      \\
      \midrule
      Total      &           & 7516      & 6198      & 6284      \\
      \bottomrule
    \end{tabular}
    \caption{Loss = {\sr}, Weighting = yes (FS=76.06, LA=78.29)}
  \end{subcaptionblock} 
  \caption{
    Confusion matrices on the development set, with the BERT Base backbone.
The ``Total'' row shows the number of times each class is predicted.
  }
  \label{tab:confusion-matrix-bert-base}
\end{table*}

\begin{table*}[t]
  \centering
  \tablefont
  \begin{subcaptionblock}[T]{0.48\linewidth}
    \centering
    \begin{tabular}{ll rrr}
      \toprule 
            &           & \multicolumn3c{Prediction}        \\
      \cmidrule{3-5}
            &           & $\labels$ & $\labelr$ & $\labeln$ \\
      \midrule 
            & $\labels$ & 5985      & 222       & 459       \\
      Gold  & $\labelr$ & 436       & 5061      & 1169      \\
            & $\labeln$ & 1032      & 1042      & 4592      \\
      \midrule
      Total &           & 7453      & 6325      & 6220      \\
      \bottomrule
    \end{tabular}
    \caption{Loss = {\ce}, Weighting = no (FS=75.98, LA=78.20)}
  \end{subcaptionblock}
\begin{subcaptionblock}[T]{0.48\linewidth}
    \centering
    \begin{tabular}{ll rrr}
      \toprule 
            &           & \multicolumn3c{Prediction}        \\
      \cmidrule{3-5}
            &           & $\labels$ & $\labelr$ & $\labeln$ \\
      \midrule 
            & $\labels$ & 5817      & 238       & 611       \\
      Gold  & $\labelr$ & 349       & 5171      & 1146      \\
            & $\labeln$ & 854       & 1032      & 4780      \\
      \midrule
      Total &           & 7020      & 6441      & 6537      \\
      \bottomrule
    \end{tabular}
    \caption{Loss = {\ce}, Weighting = yes (FS=76.74, LA=78.85)}
  \end{subcaptionblock}
  \\[5ex]
  
  \begin{subcaptionblock}[T]{0.48\linewidth}
    \centering
    \begin{tabular}{ll rrr}
      \toprule 
            &           & \multicolumn3c{Prediction}        \\
      \cmidrule{3-5}
            &           & $\labels$ & $\labelr$ & $\labeln$ \\
      \midrule 
            & $\labels$ & 6011      & 188       & 467       \\
      Gold  & $\labelr$ & 437       & 5068      & 1161      \\
            & $\labeln$ & 1019      & 940       & 4707      \\
      \midrule
      Total &           & 7467      & 6196      & 6335      \\
      \bottomrule
    \end{tabular}
    \caption{Loss = {\ova}, Weighting = no (FS=76.78, LA=78.94)}
  \end{subcaptionblock}
\begin{subcaptionblock}[T]{0.48\linewidth}
    \centering
    \begin{tabular}{ll rrr}
      \toprule 
            &           & \multicolumn3c{Prediction}        \\
      \cmidrule{3-5}
            &           & $\labels$ & $\labelr$ & $\labeln$ \\
      \midrule 
            & $\labels$ & 5858      & 258       & 550       \\
      Gold  & $\labelr$ & 359       & 5214      & 1093      \\
            & $\labeln$ & 858       & 1112      & 4696      \\
      \midrule
      Total &           & 7075      & 6584      & 6339      \\
      \bottomrule
    \end{tabular}
    \caption{Loss = {\ova}, Weighting = yes (FS=76.74, LA=78.85)}
  \end{subcaptionblock}
  \\[5ex]
  
  \begin{subcaptionblock}[T]{0.48\linewidth}
    \centering
    \begin{tabular}{ll rrr}
      \toprule 
            &           & \multicolumn3c{Prediction}        \\
      \cmidrule{3-5}
            &           & $\labels$ & $\labelr$ & $\labeln$ \\
      \midrule 
            & $\labels$ & 5942      & 214       & 510       \\
      Gold  & $\labelr$ & 406       & 5076      & 1184      \\
            & $\labeln$ & 922       & 1028      & 4716      \\
      \midrule
      Total &           & 7270      & 6318      & 6410      \\
      \bottomrule
    \end{tabular}
    \caption{Loss = {\srn}, Weighting = no (FS=76.57, LA=78.68)}
  \end{subcaptionblock}
\begin{subcaptionblock}[T]{0.48\linewidth}
    \centering
    \begin{tabular}{ll rrr}
      \toprule 
            &           & \multicolumn3c{Prediction}        \\
      \cmidrule{3-5}
            &           & $\labels$ & $\labelr$ & $\labeln$ \\
      \midrule 
            & $\labels$ & 5806      & 246       & 614       \\
      Gold  & $\labelr$ & 323       & 5148      & 1195      \\
            & $\labeln$ & 852       & 1004      & 4810      \\
      \midrule
      Total &           & 6981      & 6398      & 6619      \\
      \bottomrule
    \end{tabular}
    \caption{Loss = {\srn}, Weighting = yes (FS=76.71, LA=78.83)}
  \end{subcaptionblock}
  \\[5ex]
  
  \begin{subcaptionblock}[T]{0.48\linewidth}
    \centering
    \begin{tabular}{ll rrr}
      \toprule 
            &           & \multicolumn3c{Prediction}        \\
      \cmidrule{3-5}
            &           & $\labels$ & $\labelr$ & $\labeln$ \\
      \midrule 
            & $\labels$ & 6024      & 165       & 477       \\
      Gold  & $\labelr$ & 411       & 4989      & 1266      \\
            & $\labeln$ & 1007      & 869       & 4790      \\
      \midrule
      Total &           & 7442      & 6023      & 6533      \\
      \bottomrule
    \end{tabular}
    \caption{Loss = {\sr}, Weighting = no (FS=76.86, LA=79.02)}
  \end{subcaptionblock}
\begin{subcaptionblock}[T]{0.48\linewidth}
    \centering
    \begin{tabular}{ll rrr}
      \toprule 
            &           & \multicolumn3c{Prediction}        \\
      \cmidrule{3-5}
            &           & $\labels$ & $\labelr$ & $\labeln$ \\
      \midrule 
            & $\labels$ & 5938      & 187       & 541       \\
      Gold  & $\labelr$ & 397       & 5087      & 1182      \\
            & $\labeln$ & 884       & 971       & 4811      \\
      \midrule
      Total &           & 7219      & 6245      & 6534      \\
      \bottomrule
    \end{tabular}
    \caption{Loss = {\sr}, Weighting = yes (FS=77.01, LA=79.19)}
  \end{subcaptionblock} 
  \caption{
    Confusion matrices on the development set, with the BERT Large backbone.
}
  \label{tab:confusion-matrix-bert-large}
\end{table*}

\begin{table*}[t]
  \centering
  \tablefont
  \begin{subcaptionblock}[T]{0.48\linewidth}
    \centering
    \begin{tabular}{ll rrr}
      \toprule 
            &           & \multicolumn3c{Prediction}        \\
      \cmidrule{3-5}
            &           & $\labels$ & $\labelr$ & $\labeln$ \\
      \midrule 
            & $\labels$ & 6073      & 153       & 440       \\
      Gold  & $\labelr$ & 357       & 5127      & 1182      \\
            & $\labeln$ & 964       & 865       & 4837      \\
      \midrule
      Total &           & 7394      & 6145      & 6459      \\
      \bottomrule
    \end{tabular}
    \caption{Loss = {\ce}, Weighting = no (FS=78.03, LA=80.19)}
  \end{subcaptionblock}
\begin{subcaptionblock}[T]{0.48\linewidth}
    \centering
    \begin{tabular}{ll rrr}
      \toprule 
            &           & \multicolumn3c{Prediction}        \\
      \cmidrule{3-5}
            &           & $\labels$ & $\labelr$ & $\labeln$ \\
      \midrule 
            & $\labels$ & 5783      & 220       & 663       \\
      Gold  & $\labelr$ & 238       & 5291      & 1137      \\
            & $\labeln$ & 693       & 938       & 5035      \\
      \midrule
      Total &           & 6714      & 6449      & 6835      \\
      \bottomrule
    \end{tabular}
    \caption{Loss = {\ce}, Weighting = yes (FS=78.54, LA=80.55)}
  \end{subcaptionblock} 
                                                            \\[5ex]
  
  \begin{subcaptionblock}[T]{0.48\linewidth}
    \centering
    \begin{tabular}{ll rrr}
      \toprule 
            &           & \multicolumn3c{Prediction}        \\
      \cmidrule{3-5}
            &           & $\labels$ & $\labelr$ & $\labeln$ \\
      \midrule 
            & $\labels$ & 6032      & 148       & 486       \\
      Gold  & $\labelr$ & 321       & 5092      & 1253      \\
            & $\labeln$ & 913       & 878       & 4875      \\
      \midrule
      Total &           & 7266      & 6118      & 6614      \\
      \bottomrule
    \end{tabular}
    \caption{Loss = {\ova}, Weighting = no (FS=77.88, LA=80.00)}
  \end{subcaptionblock}
\begin{subcaptionblock}[T]{0.48\linewidth}
    \centering
    \begin{tabular}{ll rrr}
      \toprule 
            &           & \multicolumn3c{Prediction}        \\
      \cmidrule{3-5}
            &           & $\labels$ & $\labelr$ & $\labeln$ \\
      \midrule 
            & $\labels$ & 5995      & 159       & 512       \\
      Gold  & $\labelr$ & 299       & 5151      & 1216      \\
            & $\labeln$ & 826       & 864       & 4976      \\
      \midrule
      Total &           & 7120      & 6174      & 6704      \\
      \bottomrule
    \end{tabular}
    \caption{Loss = {\ova}, Weighting = yes (FS=78.55, LA=80.62)}
  \end{subcaptionblock}
                                                            \\[5ex]

  \begin{subcaptionblock}[T]{0.48\linewidth}
    \centering
    \begin{tabular}{ll rrr}
      \toprule 
            &           & \multicolumn3c{Prediction}        \\
      \cmidrule{3-5}
            &           & $\labels$ & $\labelr$ & $\labeln$ \\
      \midrule 
            & $\labels$ & 6117      & 129       & 420       \\
      Gold  & $\labelr$ & 361       & 4996      & 1309      \\
            & $\labeln$ & 962       & 771       & 4933      \\
      \midrule
      Total &           & 7440      & 5896      & 6662      \\
      \bottomrule
    \end{tabular}
    \caption{Loss = {\srn}, Weighting = no (FS=78.18 LA=80.24)}
  \end{subcaptionblock}
\begin{subcaptionblock}[T]{0.48\linewidth}
    \centering
    \begin{tabular}{ll rrr}
      \toprule 
            &           & \multicolumn3c{Prediction}        \\
      \cmidrule{3-5}
            &           & $\labels$ & $\labelr$ & $\labeln$ \\
      \midrule 
            & $\labels$ & 5913      & 227       & 526       \\
      Gold  & $\labelr$ & 275       & 5410      & 981       \\
            & $\labeln$ & 780       & 1064      & 4822      \\
      \midrule
      Total &           & 6968      & 6701      & 6329      \\
      \bottomrule
    \end{tabular}
    \caption{Loss = {\srn}, Weighting = yes (FS=78.56, LA=80.73)}
  \end{subcaptionblock}
                                                            \\[5ex]
  
  \begin{subcaptionblock}[T]{0.48\linewidth}
    \centering
    \begin{tabular}{ll rrr}
      \toprule 
            &           & \multicolumn3c{Prediction}        \\
      \cmidrule{3-5}
            &           & $\labels$ & $\labelr$ & $\labeln$ \\
      \midrule 
            & $\labels$ & 6072      & 162       & 432       \\
      Gold  & $\labelr$ & 314       & 5239      & 1113      \\
            & $\labeln$ & 915       & 981       & 4770      \\
      \midrule
      Total &           & 7301      & 6382      & 6315      \\
      \bottomrule
    \end{tabular}
    \caption{Loss = {\sr}, Weighting = no (FS=78.19, LA=80.41)}
  \end{subcaptionblock}
\begin{subcaptionblock}[T]{0.48\linewidth}
    \centering
    \begin{tabular}{ll rrr}
      \toprule 
            &           & \multicolumn3c{Prediction}        \\
      \cmidrule{3-5}
            &           & $\labels$ & $\labelr$ & $\labeln$ \\
      \midrule 
            & $\labels$ & 5901      & 213       & 552       \\
      Gold  & $\labelr$ & 237       & 5238      & 1191      \\
            & $\labeln$ & 766       & 901       & 4999      \\
      \midrule
      Total &           & 6904      & 6352      & 6742      \\
      \bottomrule
    \end{tabular}
    \caption{Loss = {\sr}, Weighting = yes (FS=78.63, LA=80.70)}
  \end{subcaptionblock} 
  \caption{
    Confusion matrices on the development set, with the RoBERTa Large backbone.
}
  \label{tab:confusion-matrix-roberta-large}
\end{table*}

To provide a comprehensive view of the compared prediction models, the confusion matrices of their predictions are presented in 
Tables~\ref{tab:confusion-matrix-bert-base}--\ref{tab:confusion-matrix-roberta-large}. We observe that the sample weighting mitigates the imbalance bias in most cases. Specifically, weighting decreases the number of predictions for the majority class ($\labels$),
for example, from 7497 to 7211 in the case of the BERT Base backbone; compare Table~\ref{tab:confusion-matrix-bert-base}(a) and~(b).

\subsection{Effect of $\lambda$} \label{sec:lambda-bce}

\begin{table*}[t]
\tablefont
\centering
\begin{tabular}{l ll ll}
\toprule 
Backbone      & Loss                       & Weighting              & LA             & FS             \\
\midrule 
BERT Base     & {\ova} ($\lambda=0.125$)   & yes ($\beta=0.999999$) & \textbf{78.13} & \textbf{76.06} \\
& \ova ($\lambda=1$)         & yes ($\beta=0.99999$)  & 77.96          & 75.91          \\
\midrule
BERT Large    & {\ova} ($\lambda=0.03125$) & yes ($\beta=0.999999$) & \textbf{78.85} & \textbf{76.74} \\
& \ova ($\lambda=1$)         & yes ($\beta=0.999999$) & 78.68          & 76.56          \\
\midrule
RoBERTa Large & {\ova} ($\lambda=0.0625$)  & yes ($\beta=0.999999$) & \textbf{80.62} & \textbf{78.55} \\
& \ova ($\lambda=1$)         & yes ($\beta=0.99999$)  & 80.05          & 77.97          \\
\bottomrule
\end{tabular}
\caption{
  Effect of tuning $\lambda$ in the \ova objective. }
\label{tab:res-ablation-bce}
\end{table*}

We introduced in the \ova objective of Eq.~\eqref{eq:ovr-objective} a hyperparameter $\lambda$ to balance the primary and auxiliary terms in the objective.

To evaluate the efficacy of calibrating the $\lambda$ parameter, we specifically examine the performance for fixed $\lambda=1$ (i.e., direct application of original \ova loss),
and that of $\lambda$ tuned over the development set.
Table~\ref{tab:res-ablation-bce} shows the results.
We note that the scores of $\lambda=1$ are considerably lower than those achieved when $\lambda$ is optimized on the development set.

\section{License of the Assets}

The FEVER 2018 dataset\footnote{\url{https://fever.ai/dataset/fever.html}} is licensed under the CC BY-SA 3.0.
The KGAT implementation\footnote{\url{https://github.com/thunlp/KernelGAT}}  is licensed under the MIT License.


\begin{thebibliography}{23}
\expandafter\ifx\csname natexlab\endcsname\relax\def\natexlab#1{#1}\fi

\bibitem[{Aly et~al.(2021{\natexlab{a}})Aly, Guo, Schlichtkrull, Thorne,
  Vlachos, Christodoulopoulos, Cocarascu, and Mittal}]{aly-etal-2021-fact}
Rami Aly, Zhijiang Guo, Michael~Sejr Schlichtkrull, James Thorne, Andreas
  Vlachos, Christos Christodoulopoulos, Oana Cocarascu, and Arpit Mittal.
  2021{\natexlab{a}}.
\newblock \href {https://doi.org/10.18653/v1/2021.fever-1.1} {The fact
  extraction and {VER}ification over unstructured and structured information
  ({FEVEROUS}) shared task}.
\newblock In \emph{Proceedings of the Fourth Workshop on Fact Extraction and
  VERification (FEVER)}, pages 1--13, Dominican Republic. Association for
  Computational Linguistics.

\bibitem[{Aly et~al.(2021{\natexlab{b}})Aly, Guo, Schlichtkrull, Thorne,
  Vlachos, Christodoulopoulos, Cocarascu, and Mittal}]{aly2021feverous}
Rami Aly, Zhijiang Guo, Michael~Sejr Schlichtkrull, James Thorne, Andreas
  Vlachos, Christos Christodoulopoulos, Oana Cocarascu, and Arpit Mittal.
  2021{\natexlab{b}}.
\newblock \href {https://openreview.net/forum?id=h-flVCIlstW} {{FEVEROUS}: Fact
  extraction and {VER}ification over unstructured and structured information}.
\newblock In \emph{Proceedings of the 35th Conference on Neural Information
  Processing Systems, Datasets and Benchmarks Track (Round 1)}.

\bibitem[{Baum and Wilczek(1988)}]{Baum1988supervised}
Eric~B. Baum and Frank Wilczek. 1988.
\newblock \href
  {https://proceedings.neurips.cc/paper/1987/hash/eccbc87e4b5ce2fe28308fd9f2a7baf3-Abstract.html}
  {Supervised learning of probability distributions by neural networks.}
\newblock In \emph{Neural Information Processing Systems}. American Institute
  of Physics.

\bibitem[{Chawla et~al.(2002)Chawla, Bowyer, Hall, and
  Kegelmeyer}]{chawla2002smote}
Nitesh~V. Chawla, Kevin~W. Bowyer, Lawrence~O. Hall, and W.~Philip Kegelmeyer.
  2002.
\newblock \href {https://doi.org/10.1613/jair.953} {{SMOTE}: Synthetic minority
  over-sampling technique}.
\newblock \emph{Journal of Artificial Intelligence Research}, 16:321--357.

\bibitem[{Cui et~al.(2019)Cui, Jia, Lin, Song, and Belongie}]{cui2019class}
Yin Cui, Menglin Jia, Tsung-Yi Lin, Yang Song, and Serge Belongie. 2019.
\newblock \href {https://doi.org/10.1109/CVPR.2019.00949} {Class-balanced loss
  based on effective number of samples}.
\newblock In \emph{Proceedings of the IEEE/CVF Conference on Computer Vision
  and Pattern Recognition}, pages 9268--9277.

\bibitem[{DeHaven and Scott(2023)}]{dehaven-scott-2023-bevers}
Mitchell DeHaven and Stephen Scott. 2023.
\newblock \href {https://doi.org/10.18653/v1/2023.fever-1.6} {{BEVERS}: A
  general, simple, and performant framework for automatic fact verification}.
\newblock In \emph{Proceedings of the Sixth Fact Extraction and VERification
  Workshop (FEVER)}, pages 58--65, Dubrovnik, Croatia. Association for
  Computational Linguistics.

\bibitem[{Devlin et~al.(2019)Devlin, Chang, Lee, and
  Toutanova}]{devlin-etal-2019-bert}
Jacob Devlin, Ming-Wei Chang, Kenton Lee, and Kristina Toutanova. 2019.
\newblock \href {https://doi.org/10.18653/v1/N19-1423} {{BERT}: Pre-training of
  deep bidirectional transformers for language understanding}.
\newblock In \emph{Proceedings of the 2019 Conference of the North {A}merican
  Chapter of the Association for Computational Linguistics: Human Language
  Technologies, Volume 1 (Long and Short Papers)}, pages 4171--4186,
  Minneapolis, Minnesota. Association for Computational Linguistics.

\bibitem[{Ishida et~al.(2017)Ishida, Niu, Hu, and
  Sugiyama}]{ishida2017complementary}
Takashi Ishida, Gang Niu, Weihua Hu, and Masashi Sugiyama. 2017.
\newblock \href
  {https://papers.nips.cc/paper_files/paper/2017/hash/1dba5eed8838571e1c80af145184e515-Abstract.html}
  {Learning from complementary labels}.
\newblock In \emph{Advances in Neural Information Processing Systems}.

\bibitem[{Ishida et~al.(2019)Ishida, Niu, Menon, and
  Sugiyama}]{pmlr-v97-ishida19a}
Takashi Ishida, Gang Niu, Aditya Menon, and Masashi Sugiyama. 2019.
\newblock \href {https://proceedings.mlr.press/v97/ishida19a.html}
  {Complementary-label learning for arbitrary losses and models}.
\newblock In \emph{Proceedings of the 36th International Conference on Machine
  Learning}, pages 2971--2980.

\bibitem[{Jiang et~al.(2021)Jiang, Pradeep, and
  Lin}]{jiang-etal-2021-exploring-listwise}
Kelvin Jiang, Ronak Pradeep, and Jimmy Lin. 2021.
\newblock \href {https://doi.org/10.18653/v1/2021.acl-short.51} {Exploring
  listwise evidence reasoning with t5 for fact verification}.
\newblock In \emph{Proceedings of the 59th Annual Meeting of the Association
  for Computational Linguistics and the 11th International Joint Conference on
  Natural Language Processing (Volume 2: Short Papers)}, pages 402--410,
  Online. Association for Computational Linguistics.

\bibitem[{Krishna et~al.(2022)Krishna, Riedel, and
  Vlachos}]{krishna-etal-2022-proofver}
Amrith Krishna, Sebastian Riedel, and Andreas Vlachos. 2022.
\newblock \href {https://doi.org/10.1162/tacl_a_00503} {{P}roo{FV}er: Natural
  logic theorem proving for fact verification}.
\newblock \emph{Transactions of the Association for Computational Linguistics},
  10:1013--1030.

\bibitem[{Liu et~al.(2019)Liu, Ott, Goyal, Du, Joshi, Chen, Levy, Lewis,
  Zettlemoyer, and Stoyanov}]{liu2019roberta}
Yinhan Liu, Myle Ott, Naman Goyal, Jingfei Du, Mandar Joshi, Danqi Chen, Omer
  Levy, Mike Lewis, Luke Zettlemoyer, and Veselin Stoyanov. 2019.
\newblock \href {https://doi.org/10.48550/arXiv.1907.11692} {{RoBERTa}: A
  robustly optimized {BERT} pretraining approach}.
\newblock {a}rXiv preprint 1907.11692 [cs.CL].

\bibitem[{Liu et~al.(2020)Liu, Xiong, Sun, and Liu}]{liu-etal-2020-fine}
Zhenghao Liu, Chenyan Xiong, Maosong Sun, and Zhiyuan Liu. 2020.
\newblock \href {https://doi.org/10.18653/v1/2020.acl-main.655} {Fine-grained
  fact verification with kernel graph attention network}.
\newblock In \emph{Proceedings of the 58th Annual Meeting of the Association
  for Computational Linguistics}, pages 7342--7351, Online. Association for
  Computational Linguistics.

\bibitem[{Mc{N}emar(1947)}]{Mcnemar1947NoteOT}
Quinn Mc{N}emar. 1947.
\newblock \href {https://doi.org/10.1007/BF02295996} {Note on the sampling
  error of the difference between correlated proportions or percentages}.
\newblock \emph{Psychometrika}, 12:153--157.

\bibitem[{Ren et~al.(2018)Ren, Zeng, Yang, and Urtasun}]{ren2018learning}
Mengye Ren, Wenyuan Zeng, Bin Yang, and Raquel Urtasun. 2018.
\newblock \href {https://proceedings.mlr.press/v80/ren18a.html} {Learning to
  reweight examples for robust deep learning}.
\newblock In \emph{Proceedings of the 35th International Conference on Machine
  Learning}, pages 4334--4343.

\bibitem[{Stammbach(2021)}]{stammbach-2021-evidence}
Dominik Stammbach. 2021.
\newblock \href {https://doi.org/10.18653/v1/2021.fever-1.2} {Evidence
  selection as a token-level prediction task}.
\newblock In \emph{Proceedings of the Fourth Workshop on Fact Extraction and
  VERification (FEVER)}, pages 14--20, Dominican Republic. Association for
  Computational Linguistics.

\bibitem[{Thorne et~al.(2018)Thorne, Vlachos, Christodoulopoulos, and
  Mittal}]{thorne-etal-2018-fever}
James Thorne, Andreas Vlachos, Christos Christodoulopoulos, and Arpit Mittal.
  2018.
\newblock \href {https://doi.org/10.18653/v1/N18-1074} {{FEVER}: a large-scale
  dataset for fact extraction and {VER}ification}.
\newblock In \emph{Proceedings of the 2018 Conference of the North {A}merican
  Chapter of the Association for Computational Linguistics: Human Language
  Technologies, Volume 1 (Long Papers)}, pages 809--819, New Orleans,
  Louisiana. Association for Computational Linguistics.

\bibitem[{Thorne et~al.(2019)Thorne, Vlachos, Cocarascu, Christodoulopoulos,
  and Mittal}]{thorne-etal-2019-fever2}
James Thorne, Andreas Vlachos, Oana Cocarascu, Christos Christodoulopoulos, and
  Arpit Mittal. 2019.
\newblock \href {https://doi.org/10.18653/v1/D19-6601} {The {FEVER}2.0 shared
  task}.
\newblock In \emph{Proceedings of the Second Workshop on Fact Extraction and
  VERification (FEVER)}, pages 1--6, Hong Kong, China. Association for
  Computational Linguistics.

\bibitem[{Tymoshenko and Moschitti(2021)}]{tymoshenko-moschitti-2021-strong}
Kateryna Tymoshenko and Alessandro Moschitti. 2021.
\newblock \href {https://doi.org/10.18653/v1/2021.findings-acl.426} {Strong and
  light baseline models for fact-checking joint inference}.
\newblock In \emph{Findings of the Association for Computational Linguistics:
  ACL-IJCNLP 2021}, pages 4824--4830, Online. Association for Computational
  Linguistics.

\bibitem[{Yu et~al.(2018)Yu, Liu, Gong, and Tao}]{yu2018learning}
Xiyu Yu, Tongliang Liu, Mingming Gong, and Dacheng Tao. 2018.
\newblock \href {https://link.springer.com/chapter/10.1007/978-3-030-01246-5_5}
  {Learning with biased complementary labels}.
\newblock In \emph{Proceedings of the European Conference on Computer Vision},
  pages 68--83.

\bibitem[{Zhang(2004)}]{zhang2004statistical}
Tong Zhang. 2004.
\newblock \href {https://www.jmlr.org/papers/v5/zhang04b.html} {Statistical
  analysis of some multi-category large margin classification methods}.
\newblock \emph{Journal of Machine Learning Research}, 5:1225--1251.

\bibitem[{Zhang et~al.(2023)Zhang, Kang, Hooi, Yan, and Feng}]{10105457}
Yifan Zhang, Bingyi Kang, Bryan Hooi, Shuicheng Yan, and Jiashi Feng. 2023.
\newblock \href {https://doi.org/10.1109/TPAMI.2023.3268118} {Deep long-tailed
  learning: A survey}.
\newblock \emph{IEEE Transactions on Pattern Analysis and Machine
  Intelligence}, 45(9):10795--10816.

\bibitem[{Zhou et~al.(2019)Zhou, Han, Yang, Liu, Wang, Li, and
  Sun}]{zhou-etal-2019-gear}
Jie Zhou, Xu~Han, Cheng Yang, Zhiyuan Liu, Lifeng Wang, Changcheng Li, and
  Maosong Sun. 2019.
\newblock \href {https://doi.org/10.18653/v1/P19-1085} {{GEAR}: Graph-based
  evidence aggregating and reasoning for fact verification}.
\newblock In \emph{Proceedings of the 57th Annual Meeting of the Association
  for Computational Linguistics}, pages 892--901, Florence, Italy. Association
  for Computational Linguistics.

\end{thebibliography}
\end{document}